\documentclass[10pt,twocolumn,letterpaper]{article}
\usepackage{wacv} 
\usepackage{graphicx}
\usepackage{amsmath}
\usepackage{amssymb}
\usepackage{booktabs}
\usepackage{tabularx}
\usepackage{shellesc}

\usepackage[pagebackref,breaklinks,colorlinks,bookmarks=false]{hyperref}

\usepackage[capitalize]{cleveref}
\crefname{section}{Sec.}{Secs.}
\Crefname{section}{Section}{Sections}
\Crefname{table}{Table}{Tables}
\crefname{table}{Tab.}{Tabs.}
\def\wacvPaperID{15} 
\def\confName{WACV}
\def\confYear{2024}
\begin{document}

\title{Cross-modal Contrastive Learning with Asymmetric Co-attention Network for Video Moment Retrieval}
\author{Love Panta$^{12}$\thanks{Both authors contributed equally to this work}, Prashant Shrestha$^{1}\footnotemark[1]$, Brabeem Sapkota$^{1}$, Amrita Bhattarai$^{1}$, \\Suresh Manandhar$^{2}$, and Anand Kumar Sah$^{1}$\\
$^{1}$IOE, Pulchowk Campus, $^{2}$Wiseyak Solutions Pvt. Ltd.\\
{\tt\small $\{$075bei016.love, 075bei024.prashant, 075bei011.brabeem, 075bei006.amrita,} \\ \tt\small {anand.sah$\}$@pcampus.edu.np, suresh.manandhar@wiseyak.com}}

\maketitle
\begin{abstract}
Video moment retrieval is a challenging task requiring fine-grained interactions between video and text modalities. Recent work in image-text pretraining has demonstrated that most existing pretrained models suffer from information asymmetry due to the difference in length between visual and textual sequences. We question whether the same problem also exists in the video-text domain with an auxiliary need to preserve both spatial and temporal information. Thus, we evaluate a recently proposed solution involving the addition of an asymmetric co-attention network for video grounding tasks. Additionally, we incorporate momentum contrastive loss for robust, discriminative representation learning in both modalities. We note that the integration of these supplementary modules yields better performance compared to state-of-the-art models on the TACoS dataset and comparable results on ActivityNet Captions, all while utilizing significantly fewer parameters with respect to baseline.
\end{abstract}

\section{Introduction}
Recent trends in machine learning have shown a growing interest in multimodal learning, specifically in vision language tasks such as visual question answering, image-text retrieval, video grounding and so on. Video moment retrieval, also known as video grounding, aims to align a video segment semantically with a given sentence query. Numerous approaches have been proposed to address video grounding, but their results were unsatisfactory due to limitations in capturing both spatial and temporal information \cite{lan2023survey}. Transformer-based methods have dominated the vision-language landscape in recent years and have also been effectively used for video grounding \cite{zhang2021multi,chen2021end,woo2022explore,yu2021cross,zhang2021video}. One advantage of using transformers over other neural network architectures is their ability to model long sequences without losing context \cite{vaswani2017attention}
and little need for engineering fusion approaches for effective multimodal interaction.

We opted for a single-stream transformer backbone due to their efficiency and little need for engineering on cross-modal interactions compared to dual-stream architectures. However, the effectiveness of single-stream multimodal architectures is limited by the imbalance in the length of visual and textual query in image-text pretraining which increases the learning time and reduces the performance of the model \cite{li2022mplug}.  
To alleviate the problem, Li \textit{et al.} \cite{li2022mplug} proposes an asymmetric co-attention block at the beginning of the network and outputs the visual-aware text features. This asymmetry still exists when we move to the video grounding task as the video feature sequences are much longer than the accompanying textual feature sequences. Thus, we adapt this approach to a transformer-based architecture proposed by MSAT \cite{zhang2021multi} for video moment retrieval.

In terms of training objectives, recent methods in image text pretraining \cite{li2022mplug, li2021align, he2022vlmae} have extensively adapted contrastive loss together with transformers for effective multimodal interaction. However, few works have been proposed for video grounding \cite{zhang2021video,chen2021end}. Inspired by approaches in image text pretraining, we employ additional Video Text Contrastive (VTC) loss to our architecture. Experiments on MSAT \cite{zhang2021multi} have also shown the effectiveness of the loss. Additionally, we observe that VTC loss allows the decoupled attention paradigm of MSAT to be dropped without affecting performance, greatly reducing the model parameters. Moreover, MSAT architecture introduces the novel multi-stage aggregated module(MSA) on the top of their transformer module to capture the stage-specific information which is also integrated into our model.

In summary, our key contributions are three-fold:
\begin{itemize}
    \item We evaluate the effectiveness of a recently proposed solution to the information asymmetry problem inspired by image-text pretraining on the video grounding task.
    \item We employ the momentum contrastive loss for more robust feature learning across both visual and text modalities, thereby achieving better or comparable results on both datasets surpassing various state-of-arts.
     \item We conduct experiments to assess the effectiveness of various modules on both our architecture and the baseline, drawing conclusions about our superior performance.
\end{itemize}

\section{Related Work}
The video grounding task itself is a challenging task that requires a high-level understanding of semantic relations between video and text features \cite{yang2020survey,zhang2022temporal,lan2023survey}. Many approaches have been proposed so far which can be broadly categorized into proposal-based and proposal-free methods. 
Proposal based methods \cite{zhang2020learning,zhang2021multi,zhang2021video,chen2021end} typically use a two-stage framework. This involves either utilizing a predefined set of candidate moments or generating such candidates. These candidates are then ranked by the model based on their relevance to the provided sentence. In contrast proposal-free methods \cite{yu2021cross,mun2020local,li2021proposal,zhang2021natural} aim to directly predict moment boundaries, eliminating the need for explicit proposals. These methods either directly estimate temporal boundaries or adopt a span-based approach to assign probabilities to each video index, indicating its potential as a starting or ending point for the moment.

A key engineering concern within proposal-based methods lies in the representation of proposals. Pooling-based strategies often lack the needed discrimination for precise localization. To address this, Zhang \textit{et al.} \cite{zhang2021multi} introduces a distinctive stage-specific representation method for enhancing proposal representation. From an architectural perspective, transformers have shown their effectiveness in various multimodal learning tasks, including video grounding \cite{zhang2022temporal,zhang2021video,yu2021cross,zhang2021multi,chen2021end}. 
The attention-based mechanism in transformers enables the model to efficiently capture multi-modal relations, along with spatio-temporal context information, facilitating improved alignment between the text and video \cite{yang2022tubedetr, su2021stvgbert}.
\par
Contrastive learning aims to learn representations that maximize the similarity scores between positive pairs thereby making the encoder more discriminative. In the following papers \cite{he2020momentum,chen2020simple}, the authors introduce contrastive learning in the visual domain. Inspired by MoCo \cite{he2020momentum}, ALBEF \cite{li2021align} introduces the multi-modal contrastive loss in image-text pretraining which shows the effectiveness of the approach on various downstream tasks. For video grounding and video corpus moment retrieval tasks, the following papers \cite{nan2021interventional, zhang2021video} propose multi-modal contrastive learning which maximizes the mutual information between two modalities to learn the more robust representations in an unsupervised way. However, these approaches require large mini-batches which is computationally inefficient and results in low accuracy.

\section{Method}
Our architecture mainly consists of three main components i.e., visual language constrastive loss,
transformer backbone with asymmetric co-attention and multistage-aggregated module from \cite{zhang2021multi}. These components are trained end to end after freezing the feature extractors to predict the target moment given text query.

\subsection{Feature extraction}
Given a video $X$, it is divided into sequence of frames as $V = \{x_1, x_2, ..., x_f\}$. Then, pre-trained C3D \cite{tran2015learning} is employed to extract the spatio-temporal features on the bulk of frame sequences. The resulting feature vectors undergo mean pooling to standardize the segment count of features, ensuring a consistent value regardless of the video's duration. Finally, the video is represented as 
$V=\left\{v_{i}\right\}_{i=1}^{N}$ where, $N$ is the length of the video clip sequence chosen.
Each video has its corresponding annotation $\{T_q, ts, te\}$, where, ${T_q}$ is the textual query, $t_s$ and $t_e$ are the true starting and ending index of the video moment that corresponds to the language sentence.

For the textual query, GloVe\cite{pennington2014GloVe} embeddings are used to produce a sequence of word-level feature vectors. Thus, a sentence in a dataset is expressed as $T_q=\left\{w_{i}\right\}_{i=1}^{M}$, where, $w_{i}$ is GloVe embedding of $i^{th}$ word in a sentence $T_q$ of length $M$.

\subsection{Visual Language Transformer Block}

\begin{figure}[t!]
\includegraphics[width=1.0\columnwidth]{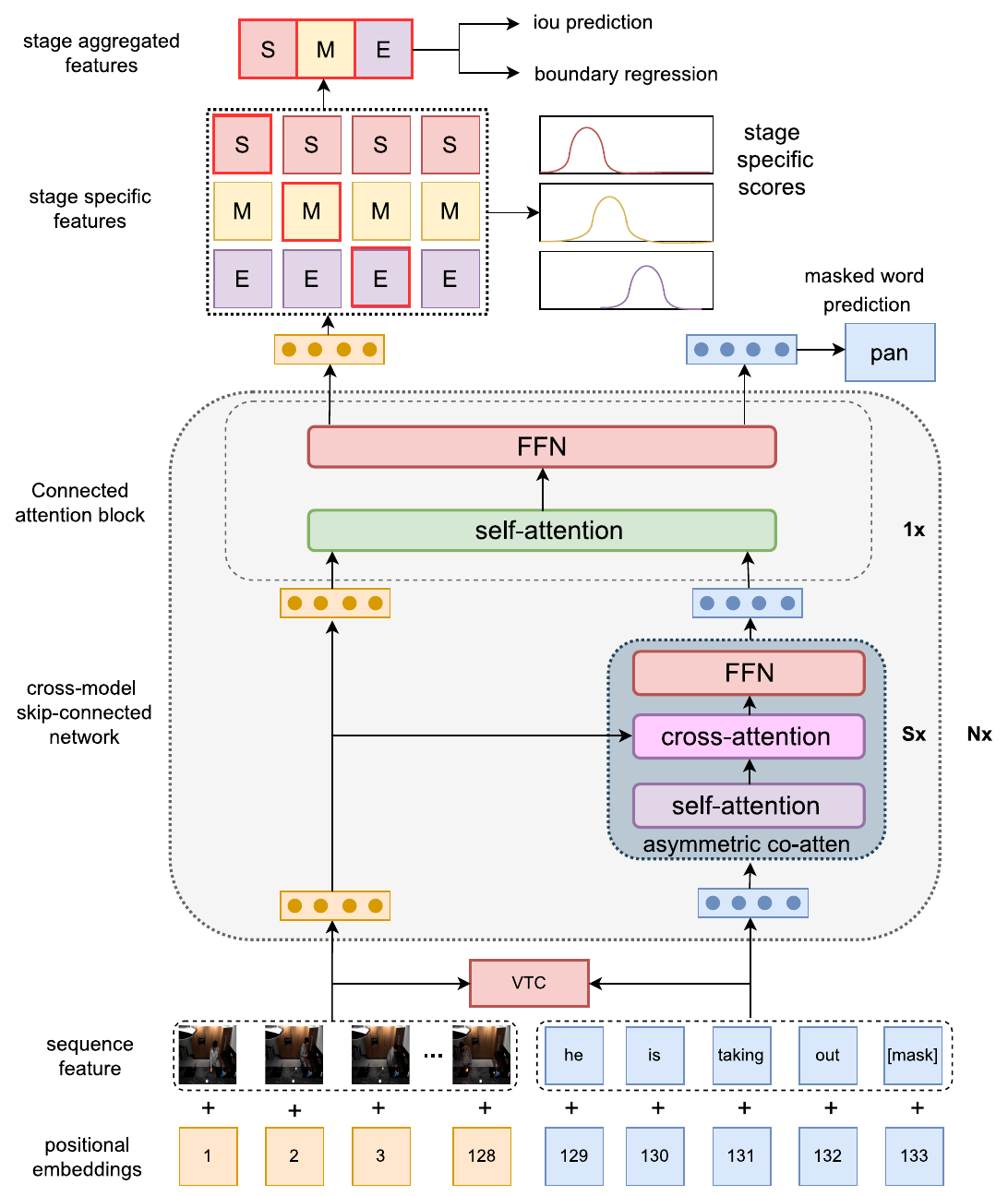}
\caption{Cross-model Skip-connected Network}
\label{mplug_reformed}
\end{figure}
Following the work of mPLUG \cite{li2022mplug}, we propose to use a similar transformer architecture for better vision-language understanding in the context of video moment retrieval. They utilize cross-modal skip-connections, enabling fusion at disparate abstraction levels, and creating inter-layer shortcuts to capture semantic richness in language compared to vision. In the video domain, employing the same architecture is highly likely to be beneficial, given the necessity of capturing both spatial and temporal context information from the attended language features.
\par
The main transformer backbone consists of sequentially arranged $N$ cross-model skip-connected blocks. Each block is formed by the $S$ repeated asymmetric co-attention layers followed by a single connected self-attention block as shown in Fig. \ref{mplug_reformed}, where $S$ is the stride layer value.
The input for the transformer backbone comprises visual and text features obtained from the feature encoder in the contrastive learning block. Subsequently, we iterate the input sequence through the asymmetric co-attention block to acquire visual aware text features. The features thus obtained from both domains are concatenated and passed into the single self-attention block referred to as Connected Attention Block(CAB). The whole process is repeated multiple times until the semantically rich text features are given for masked language prediction task \cite{devlin2018bert} whereas on the vision side, it is passed into the multi-stage aggregated module \cite{zhang2021multi}.

\subsection{Proposal Generation and Ranking}
We follow the approach taken by \cite{zhang2021multi} and pass the encoded visual features to a multi-stage aggregated module. The candidate proposals are generated using a 2D temporal map \cite{zhang2020learning}. The multi-stage aggregated module provides temporal stage-specific representations for each video clip i.e. beginning, middle and ending stages. Then, for each moment candidate, the stage-specific features are sampled and concatenated to produce proposal representations. These stage-specific representations are more discriminative than pooling-based representations. These proposals are boundary-regressed and ranked to produce the final output. More details can be found on the paper \cite{zhang2020learning}.

We combine the losses used by \cite{zhang2021multi} together with the Contrastive loss explained in the next section to train the model.
\subsection{Contrastive Loss}
\begin{figure}[t!]
\includegraphics[width=1.0\columnwidth]{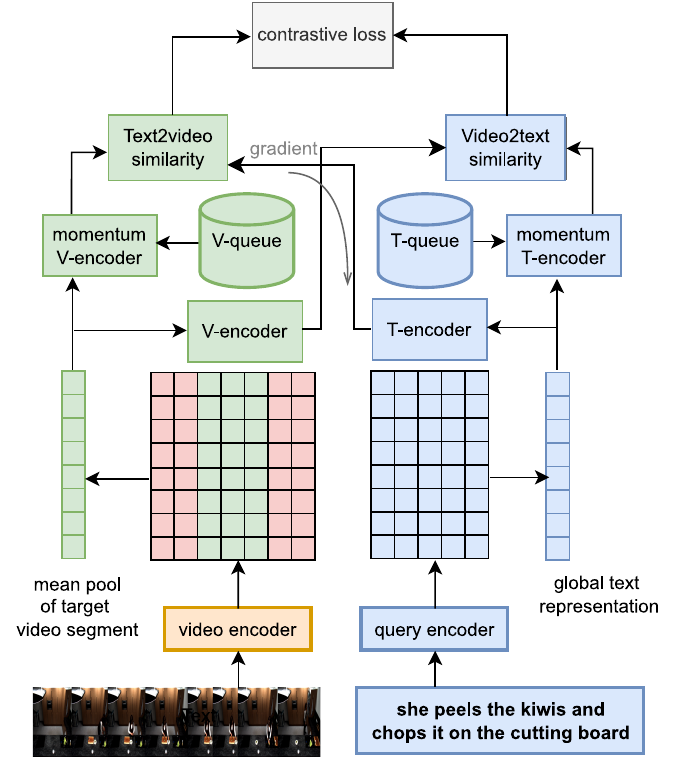}
\caption{Video-text momentum contrastive learning}
\label{VTC}
\end{figure}

Similar to Image Text Contrastive Learning loss (ITC) adopted by \cite{li2021align}, Video Text Contrastive Learning loss (VTC) is used before passing the features into the transformer backbone as shown in Fig. \ref{VTC}. The loss tries to align the features of positive pairs. The mean pooled target video segment and global text representations of the same training examples form the positive pairs while all other examples form the negative pairs.

The visual and text features obtained from the respective feature extractor are projected to the same latent dimension and then the positional embedding is added to preserve the sequence information which forms the input for the contrastive encoder block. Based on MoCo \cite{he2020momentum} and ALBEF \cite{li2021align}, a similarity function, $s=g_v\left(\boldsymbol{v}_{\mathrm{avg}}\right)^{\top} g_w\left(\boldsymbol{w}_{\mathrm{avg}}\right)$ is learnt such that video segments that align with the text query have a higher similarity score. Here, $g_v$ and $g_w$ are linear transformations that map the average pooled representation to a normalized latent space. For the video modality, we first separate the features into ground truth segments and non-ground truth segments.  
For each batch, the average pooled latent feature of video ground truth portions together with the average pooled latent feature of the aligned textual query form the positive pairs. Two queues are maintained to store the $M$ most recent video-text representations from the momentum unimodal encoders to generate the negative pairs. The normalized features from momentum encoders are denoted as $g_v^{\prime}\left(\boldsymbol{v}_{\mathrm{avg}}^{\prime}\right)$ and $g_w^{\prime}\left(\boldsymbol{w}_{\mathrm{avg}}^{\prime}\right)$.
The two-way similarity is defined by
$s(V, T)=g_v\left(\boldsymbol{v}_{\mathrm{avg}}\right)^{\top} g_w^{\prime}\left(\boldsymbol{w}_{\mathrm{avg}}^{\prime}\right)$ and $s(T, V)=g_w\left(\boldsymbol{w}_{\mathrm{avg}}\right)^{\top} g_v^{\prime}\left(\boldsymbol{v}_{\mathrm{avg}}^{\prime}\right)$.

For each video segment and text, the softmax-normalized video-to-text and text-to-video similarity is calculated as:

\begin{equation}
    \begin{cases}
    & \normalsize p_m^{\mathrm{v} 2 \mathrm{t}}(V)=\dfrac{\exp \left(s\left(V, T_m\right) / \tau\right)}{\sum_{m=1}^M \exp \left(s\left(V, T_m\right) / \tau\right)} \\
    & \normalsize p_m^{\mathrm{t} 2 \mathrm{v}}(T)=\dfrac{\exp \left(s\left(T, V_m\right) / \tau\right)}{\sum_{m=1}^M \exp \left(s\left(T, V_m\right) / \tau\right)}
    \end{cases}
\end{equation}

where $\tau$ is a temperature parameter which is learnable. If $\boldsymbol{y}^{\mathrm{v} 2 \mathrm{t}}(V)$ and $\boldsymbol{y}^{\mathrm{t} 2 \mathrm{v}}(T)$ denote the ground-truth one-hot similarity, where negative pairs have a value of $0$ and positive pairs have a value of $1$. The video-text contrastive loss is calculated as the cross entropy between $p$ and $y$:
\begin{equation}
    \begin{split}
     \mathcal{L}_{\text {vtc}}= \frac{1}{2} \mathbb{E}_{(V, T) \sim D} \Bigl[ 
     & \mathrm{H}\left(\boldsymbol{y}^{\mathrm{v} 2 \mathrm{t}}(V), \boldsymbol{p}^{\mathrm{v} 2 \mathrm{t}}(V)\right)+ 
     \\& \mathrm{H}\left(\boldsymbol{y}^{\mathrm{t} 2 \mathrm{v}}(T), \boldsymbol{p}^{\mathrm{t} 2 \mathrm{v}}(T)\right)\Bigl]
    \end{split}
\end{equation}
ALBEF address the weak correlation of positive pairs and the correlation of negative pairs by learning from pseudo-targets generated by the momentum model. This momentum distillation approach is also used for the VTC loss.

\section{Experiments}
\subsection{Datasets}
\subsubsection{TACoS}
The TACoS dataset was introduced in Regneri \textit{et al.} \cite{regneri2013grounding} and is a popular benchmark dataset used in the literature. It contains a total of 127 videos with an average duration of about 7 minutes. The train, val and test splits use the standard split of 10,146, 4,589, and 4,083 video-segment - query pairs respectively. The videos in the TACoS dataset were built on top of the “MPII Cooking Composite Activities” video corpus (Rohrbach \textit{et al.}, 2012, MPII Composites) \cite{rohrbach2012script}, containing videos of various cooking activities, e.g., cutting kiwi, cleaning chopping board, etc.
The actions performed are in the same kitchen but the lighting conditions do vary. TACoS is considered a challenging dataset because of the level of detail in some of the queries in the dataset.

~\par 
\subsubsection{ActivityNet Captions}
The ActivityNet Captions was introduced by Krishna \textit{et al.} \cite{krishna2017dense} for dense video captioning tasks. Here, there are 20k videos and 100k annotations in total, with an average of 4.82 temporally localized sentences per video. The average duration of the videos is 2 minutes.
The validation subset ”val 1” is used as the validation set while the subset ”val 2” is used as the test set. As in \cite{zhang2021multi} and \cite{zhang2020learning}, for the train, val and test set of 37,417, 17,505 and 17,031 video-segment - query pairs are used respectively. The ActivityNet Captions are characteristic in that the range of domains of the videos is vast. Thus, the results on the ActivityNet Captions may be more relevant in the context of the use of the models in general settings.
\par
One characteristic feature of the ActivityNet dataset is the decreased level of detail of the annotations. Compared to TACoS, the ground truth moments are longer while the total video duration is shorter. Thus, there is a high likelihood of a random prediction to overlap with the ground truth moment. 

\subsection{Implementation Details}

AdamW optimizer \cite{loshchilov2017decoupled} is used with a learning rate of 5$e^{-5}$ to train the model for both datasets. We set the number of transformer blocks to be 3 and the stride length for each asymmetric co-attention block to be 2. For the VTC module, a simple linear layer is used as an encoder block for both video and query features to project it into a common latent space for calculating similarity scores. For negative sample mining, a queue size of 50,000 is used and the momentum encoders are set to have the distillation weight, $\alpha$ of 0.3, momentum parameter of 0.995 for updating the respective momentum encoders, and temperature coefficient ($\tau$) of 0.1. All hyperparameters for the proposal generation, the multi-stage aggregated module and the feature extraction are set according to \cite{zhang2021multi}.

\subsection{Evaluation Metric}
For the sake of comparison, $\mathrm{R} @ n$, IoU@$m$ is used for evaluation. It refers to the percentage of text queries, for which IoU of at least one of the $n$ temporal moment predictions with the ground truth exceeds $m$. For example if one of the predictions for the query, $q_{i}$ results in an IoU with the ground truth of over $m$, then $r\left(n, m, q_{i}\right)=1$. Otherwise, $r\left(n, m, q_{i}\right)=0$. Thus, $\mathrm{R} @ n$, $\text{IoU}@m$ is calculated as:
\begin{equation}
\mathrm{R} @ n, \mathrm{IoU} @ m=\frac{1}{N_{q}} \sum_{i=1}^{N_{q}} r\left(n, m, q_{i}\right)
\end{equation}
~\par
We use $n \in \{ 1, 5\}$ for both the datasets. For the ActivityNet Captions, $m \in \{0.5, 0.7\}$ and for the TACoS dataset, $m\in \{0.3, 0.5\}$.

\subsection{Comparison with other methods}

\begin{table}[!ht]
    \caption{Comparisons with SOTA on TACoS dataset based on C3D features}
    \centering
    \renewcommand{\arraystretch}{1.3}
     \resizebox{8cm}{!} {
    \begin{tabularx}{4in}{|c|X|X|X|X|}
    \hline
        \textbf{Method} & \textbf{R@1, IoU@0.3$\uparrow$} & \textbf{R@1, IoU@0.5$\uparrow$} & \textbf{R@5, IoU@0.3$\uparrow$} & \textbf{R@5, IoU@0.5$\uparrow$} \\ \hline
        2D-TAN \cite{zhang2020learning} & 37.29 & 25.32 & 57.81 & 45.04 \\ \hline
        FIAN \cite{qu2020fine} & 33.87 & 28.58 & 47.76 & 39.16 \\ \hline
        CSMGAN \cite{liu2020jointly} & 33.9 & 27.09 & 53.98 & 41.22 \\ \hline
        IVG \cite{nan2021interventional} & 38.84 & 29.07 & - & - \\ \hline
        DPIN \cite{wang2020dual} & 46.74 & 32.92 & 62.16 & 50.26 \\ \hline
        SMIN \cite{wang2021structured} & 48.01 & 35.24 & 65.18 & 53.36 \\ \hline
        MSAT \cite{zhang2021multi} & 48.79 & 37.57 & 67.63 & 57.91 \\ \hline
        STCM-Net \cite{jia2022stcm} & 49.04 & 35.59 & \textbf{70.13} & 57.69 \\ \hline
        OURS & \textbf{49.77} & \textbf{37.99} & 68.31 & \textbf{58.31} \\ \hline
    \end{tabularx}}
    \label{tacos_eval}
\end{table}
\begin{table}[!ht]
     \caption{Comparisons with SOTA on ActivityNet Captions dataset based on C3D features}
    \centering
     \renewcommand{\arraystretch}{1.3}
     \resizebox{8cm}{!} {
    \begin{tabularx}{4in}{|c|X|X|X|X|}
    \hline
        \textbf{Method} & \textbf{R@1, IoU@0.5$\uparrow$} & \textbf{R@1, IoU@0.7$\uparrow$} & \textbf{R@5, IoU@0.5$\uparrow$} & \textbf{R@5, IoU@0.7$\uparrow$} \\ \hline
        2D-TAN \cite{zhang2020learning} & 44.51 & 26.54 & 77.13 & 61.96 \\ \hline
        FIAN \cite{qu2020fine} & 47.9 & 29.81 & 77.64 & 59.66 \\ \hline
        CSMGAN \cite{liu2020jointly} & \textbf{49.11} & 27.09 & 77.43 & 59.63 \\ \hline
        IVG \cite{nan2021interventional} & 43.84 & 27.10 & - & - \\ \hline
        DPIN \cite{wang2020dual} & 47.27 & 28.31 & 77.45 & 60.03 \\ \hline
        SMIN \cite{wang2021structured} & 48.46 & 30.34 & \textbf{81.16} & 62.11 \\ \hline
        MSAT \cite{zhang2021multi} & 48.02 & \textbf{31.78} & 78.02 & 63.18 \\ \hline
        STCM-Net \cite{jia2022stcm} & 46.23 & 29.04 & 78.43 & 63.46 \\ \hline
        OURS & 47.73 & 31.21 & 78.06 & \textbf{63.63} \\ \hline
    \end{tabularx}}
    \label{activitynet_eval}
\end{table}

We compare our approach with previous state-of-the-art methods and the results are shown in Table \ref{tacos_eval} and Table \ref{activitynet_eval}. Our method performs strongly against the various baselines. In TACoS, our method outperforms the baselines in almost all the metrics whereas in ActivityNet captions, it achieves comparable performances. Compared with MSAT \cite{zhang2021multi}, our method achieves 0.98 point improvement for $\mathrm{R} @ 1$,$\text{IoU}@0.3$ and provides small improvements over all other metrics in TACoS with considerably fewer parameters as seen in Table \ref{model_size}. The introduction of an asymmetric attention block reduces the overall model parameters by removing the need for decoupled attention weights. Our model requires 30 per cent fewer parameters as compared to MSAT for training the model.

2D-TAN \cite{zhang2020learning} uses a 2D temporal map of features to represent the moment candidates. It then uses 2D convolution operations to consider the temporal relation between adjacent video moments for discriminative localization. DPIN \cite{wang2020dual} instead uses two interacting branches for frame level and candidate level representations to make predictions consistent with both query semantics and moment boundaries. Furthermore, FIAN \cite{qu2020fine} applies the iterative cross-modal attention network to generate visual aware sentence representations and vice versa, whereas, CSMGAN \cite{liu2020jointly} utilizes the joint cross and self-modal graph attention network to capture the detailed high-level interactions.
\par
SMIN \cite{wang2021structured} considers the boundary and content level moment representations for coarse to fine-grained cross-model interactions. To obtain additional information from textual modality, STCM-Net \cite{jia2022stcm} further proposes the time concept mining network to extract time-related information from the sentence query. IVG \cite{nan2021interventional} applies causal inference to remove spurious correlation between video and query features. Additionally, they apply intermodal contrastive learning to align video and text features, and intramodal video-video contrastive learning to improve visual representation. MSAT \cite{zhang2021multi} uses decoupled attention weights inside multimodal transformer backbone and stage-specific representations for proposals.
\par
In a nutshell, even though MSAT outperformed past approaches in most metrics, our modification achieved even better overall performance with significantly fewer parameters.

\subsection{Ablation Study}
\begin{table}[!ht]
    \caption{Ablation study on ActivityNet Captions}
    \centering
    \renewcommand{\arraystretch}{1.3}
     \resizebox{8cm}{!} {
    \begin{tabularx}{4.4in}{|c|X|X|X|X|}
    \hline
        \textbf{Method} & \textbf{R@1, IoU@0.5$\uparrow$} & \textbf{R@1, IoU@0.7$\uparrow$} & \textbf{R@5, IoU@0.5$\uparrow$} & \textbf{R@5, IoU@0.7$\uparrow$} \\ 
        \hline
        De-VLTrans+MSA \cite{zhang2021multi} & \textbf{48.02} & \textbf{31.78} & 78.02 & 63.18 \\ 
        \hline
        ACB+De-VLTrans+MSA & 47.99 & 30.86 & 77.51 & 62.37 \\ 
        \hline
        CAB+MSA & 46.62 & 29.54 & 77.22 & 60.91 \\
        \hline
        ACB+CAB+MSA & 47.38 & 30.32 & 77.18 & 61.20 \\ 
        \hline
         VTC+De-VLTrans+MSA & 47.98 & 31.40 & 77.61 & 62.66 \\ 
        \hline
        \textbf{VTC+ACB+CAB+MSA} & 47.73 & 31.21 & \textbf{78.06} & \textbf{63.63} \\ 
        \hline
        \end{tabularx}}
    \label{activitynet_ablation}
\end{table}

We perform ablation on different components of the architecture to verify the effectiveness of our modifications to the original MSAT architecture. Table \ref{activitynet_ablation} and \ref{tacos_ablation} report the scores obtained with the different variants. While the initial design employs six distinct transformer encoder blocks, leading to higher parameter counts, our experimental setup involves the integration of only three primary transformer blocks whether in the form of De-VLTrans or CAB. This adjustment is attributed to the inclusion of supplementary components. Specifically, we study the results of the following variants.
\begin{table}[!ht]
     \caption{Ablation study on TACoS}
    \centering
    \renewcommand{\arraystretch}{1.3}
      \resizebox{8cm}{!} {
    \begin{tabularx}{4.4in}{|c|X|X|X|X|}
    \hline
        \textbf{Method} & \textbf{R@1, IoU@0.3$\uparrow$} & \textbf{R@1, IoU@0.5$\uparrow$} & \textbf{R@5, IoU@0.3$\uparrow$} & \textbf{R@5, IoU@0.5$\uparrow$} \\ 
        \hline
        De-VLTrans+MSA \cite{zhang2021multi} & 48.79 & 37.57 & 67.63 & 57.91 \\ 
        \hline
        ACB+De-VLTrans+MSA & 48.76 & 36.79 & 68.58 & 57.49 \\
        \hline
        CAB+MSA & 46.23 & 35.30 & 65.64 & 55.74 \\
        \hline
        ACB+CAB+MSA & 47.44 & 35.49 & \textbf{68.71} & 56.91 \\ 
        \hline
          VTC+De-VLTrans+MSA & 47.09 & 37.78 & 67.24 & 58.08 \\ 
        \hline
        \textbf{VTC+ACB+CAB+MSA} & \textbf{49.77} & \textbf{37.99} & 68.31 & \textbf{58.31} \\
        \hline
    \end{tabularx}}
    \label{tacos_ablation}
\end{table}
\begin{itemize}
    \item \textbf{De-VLTrans+MSA} entry reports the scores obtained by \cite{zhang2021multi} without any added modifications and refers to the complete MSAT architecture. 
    \item \textbf{ACB+De-VLTrans+MSA} refers to the use of Asymmetric Co-attention Blocks(ACB) before each \textbf{De-VLTrans} layer.
    \item \textbf{CAB+MSA} replaces the decoupled attention weighted self-attention layer of \textbf{De-VLTrans+MSA} with a simple connected self-attention layer.
     \item \textbf{ACB+CAB+MSA} refers to addition of  Asymmetric Co-attention Blocks infront of \textbf{CAB+MSA} model.
    \item \textbf{VTC+ACB+CAB+MSA} adds an extra Video-Text Contrastive loss and required encoder and momentum encoder layers to the \textbf{ACB+CAB+MSA}.
    \item \textbf{VTC+De-VLTrans+MSA} only adds the Video-Text Contrastive loss to the \textbf{De-VLTrans+MSA} architecture.
\end{itemize}

Table \ref{model_size} shows the number of trainable parameters of the original \textbf{De-VLTrans+MSA} and our best performing \textbf{VTC+ACB+CAB+MSA} variant. 
By comparing the results from above Table \ref{activitynet_ablation} and \ref{tacos_ablation}, we can clearly see that the VTC+ACB+CAB+MSA module outperforms all the other variants including the MSAT \cite{zhang2021multi} architecture in TACoS and obtains comparable results in ActivityNet. 

\begin{table}[!ht]
     \caption{Number of trainable parameters}
    \centering
     \renewcommand{\arraystretch}{1.3}
     \resizebox{8cm}{!} {
    \begin{tabularx}{3.5in}{|c|X|X|}
        \hline
         \textbf{Model} & \textbf{TACoS} & \textbf{ActivityNet}\\
         \hline
         De-VLTrans+MSA \cite{zhang2021multi} & 36M & 37M\\
         \hline
         \textbf{OURS} & \textbf{22M} & \textbf{25M}\\
         \hline
    \end{tabularx}
    }
    \label{model_size}
\end{table}

Furthermore, we also evaluate the effectiveness of adding a VTC or ACB module in MSAT architecture and find that the added module doesn't improve the accuracy instead achieves comparable performances in both datasets with decreased parameter counts.

Although the addition of individual ACB or VTC components in our architecture does not improve the evaluation results, the combination of both is seen to be effective.

\begin{figure}[ht!]
\includegraphics[width=1.0\columnwidth]{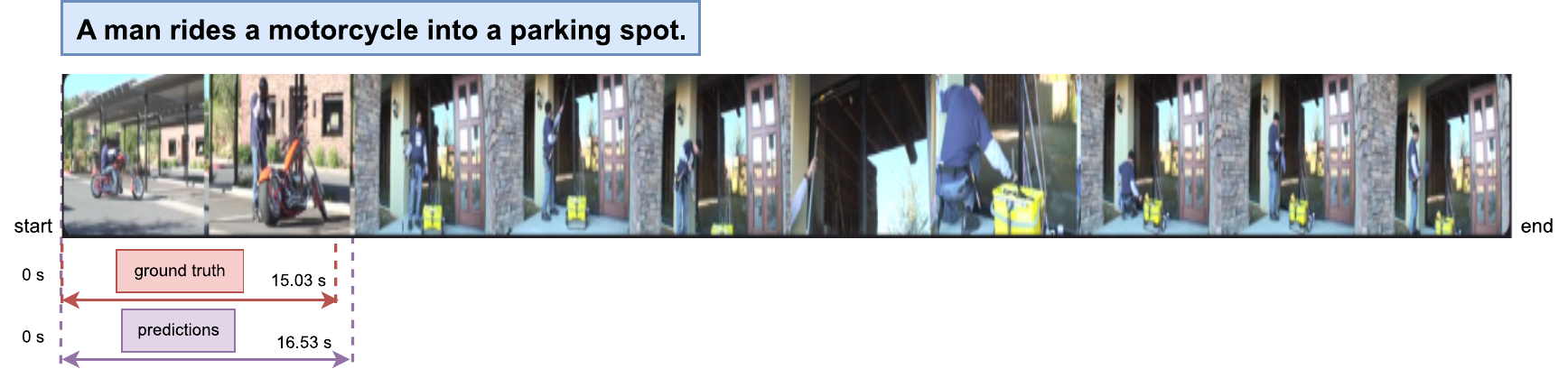}
\caption{An example of video grounding on ActivityNet dataset}
\label{activitynet_demo}
\end{figure}

\begin{figure}[ht!]
\includegraphics[width=1.0\columnwidth]{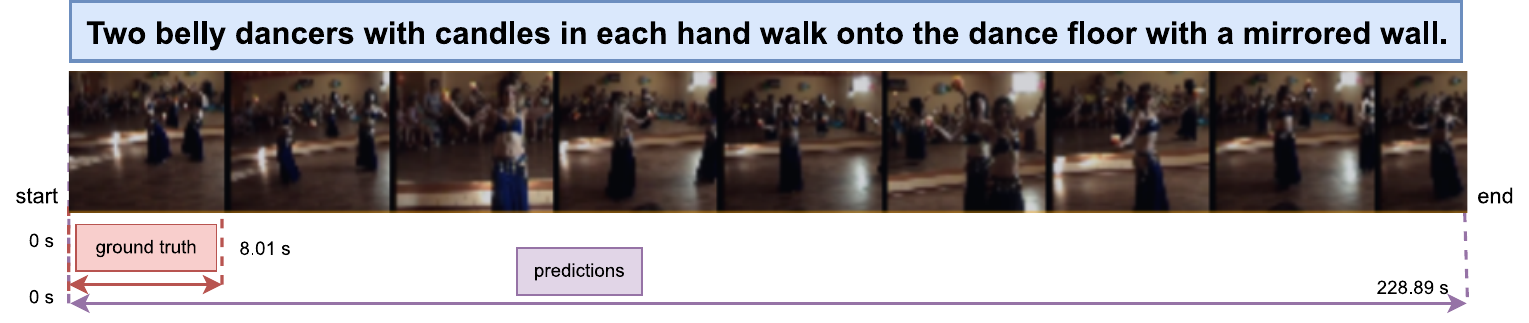}
\caption{An example of the false predictions on ActivityNet dataset}
\label{activitynet_demo_false}
\end{figure}

\subsection{Qualitative Results}
\begin{figure}[ht!]
\includegraphics[width=1.0\columnwidth]{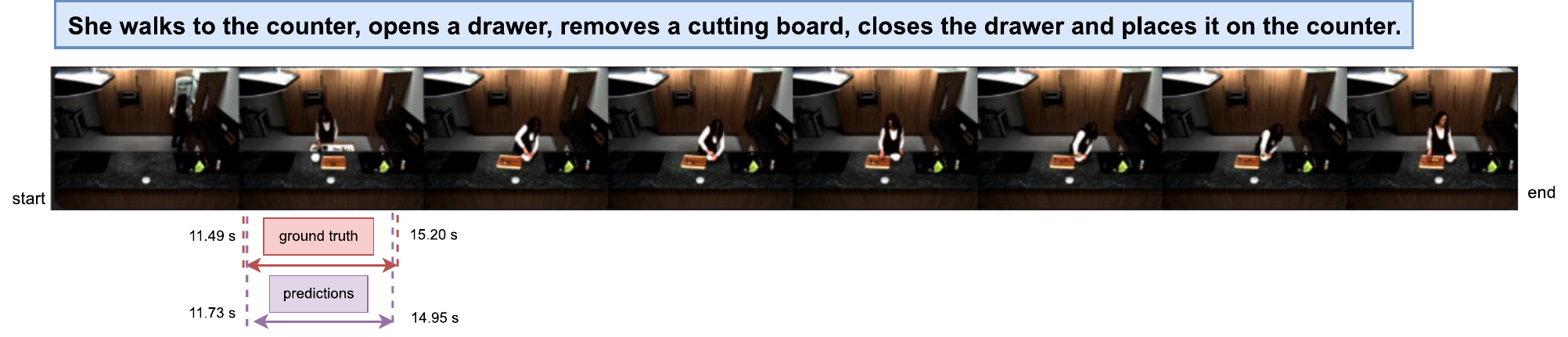}
\caption{An example of video grounding on TACoS dataset}
\label{tacos_demo}
\end{figure}

We qualitatively test both our models on the corresponding dataset as shown in Fig. \ref{activitynet_demo}, \ref{activitynet_demo_false} and \ref{tacos_demo}. In the TACoS dataset, our model effectively captures the stage-specific information and accurately localizes the temporal moments. But, in the case of the ActivityNet dataset, even if, the prediction is quite good shown in Fig. \ref{activitynet_demo}, the bias in the dataset poses a challenge to generalize to the broader domain of activity recognition tasks which can be well depicted in Fig. \ref{activitynet_demo_false}. The false temporal localization is due to the reason that our model is not able to distinguish the words "walk" and "dance" and therefore, generalize the whole video as dancing instead of recognizing the walking moment as separate actions. 

\section{Conclusion}
In this work, we evaluated the effectiveness of using asymmetric co-attention layers and video text contrastive learning in the context of video grounding. Specifically, we observed that while the addition of an asymmetric co-attention block or the contrastive loss alone does not bring any performance gain, the combined use of both modules improves over the baselines in TACoS and performs comparably in ActivityNet. Additionally, our approach requires considerably fewer learnable parameters and captures more robust multi-modal interactions across both modalities compared to the baseline architecture.

{\small
\bibliographystyle{ieee_fullname}
\bibliography{references}
}
\end{document}